\documentclass[10pt,twocolumn,letterpaper]{article}

\usepackage{cvpr}
\usepackage{times}
\usepackage{epsfig}
\usepackage{graphicx}
\usepackage{amsmath,amssymb,bm}

\usepackage[pagebackref=false,breaklinks=true,letterpaper=true,colorlinks,bookmarks=false]{hyperref}
\cvprfinalcopy 

\def\cvprPaperID{399} 

\ifcvprfinal\fi

\newcommand{\datasetName}{CompCars}

\begin{document}

\title{A Large-Scale Car Dataset for Fine-Grained Categorization and Verification}

\author{Linjie Yang$^{1}$~~~~~~~~~~~~Ping Luo$^{2,1}$~~~~~~~~~~~~Chen Change Loy$^{1,2}$~~~~~~~~~~~~Xiaoou Tang$^{1,2}$\\
$^1$Department of Information Engineering, The Chinese University of Hong Kong\\
$^2$Shenzhen Key Lab of CVPR, Shenzhen Institutes of Advanced Technology, \\
Chinese Academy of Sciences, Shenzhen, China\\
{\tt\small \{yl012,pluo,ccloy,xtang\}@ie.cuhk.edu.hk}
}

\maketitle

\begin{abstract}
This paper aims to highlight vision related tasks centered around ``car'', which has been largely neglected by vision community in comparison to other objects.
We show that there are still many interesting car-related problems and applications, which are not yet well explored and researched.
To facilitate future car-related research, in this paper we present our on-going effort in collecting a large-scale dataset, ``CompCars'', that covers not only different car views, but also their different internal and external parts, and rich attributes. Importantly, the dataset is constructed with a cross-modality nature, containing a surveillance-nature set and a web-nature set. We further demonstrate a few important applications exploiting the dataset, namely car model classification, car model verification, and attribute prediction. We also discuss specific challenges of the car-related problems and other potential applications that worth further investigations.
The latest dataset can be downloaded at \url{http://mmlab.ie.cuhk.edu.hk/datasets/comp_cars/index.html}

** Update: This technical report serves as an extension to our earlier work~\cite{yang2015large} published in CVPR 2015. The experiments shown in Sec.~\ref{sec:updated} gain better performance on all three tasks, \ie~car model classification, attribute prediction, and car model verification, thanks to more training data and better network structures. The experimental results can serve as baselines in any later research works. The settings and the train/test splits are provided on the project page.


** Update 2: This update provides preliminary experiment results for fine-grained classification on the surveillance data of CompCars. The train/test splits are provided in the updated dataset. See details in Section~\ref{sec:surveillance}.
\end{abstract}

\section{Introduction}

\begin{figure}[t]\centering
\includegraphics[width=1\linewidth]{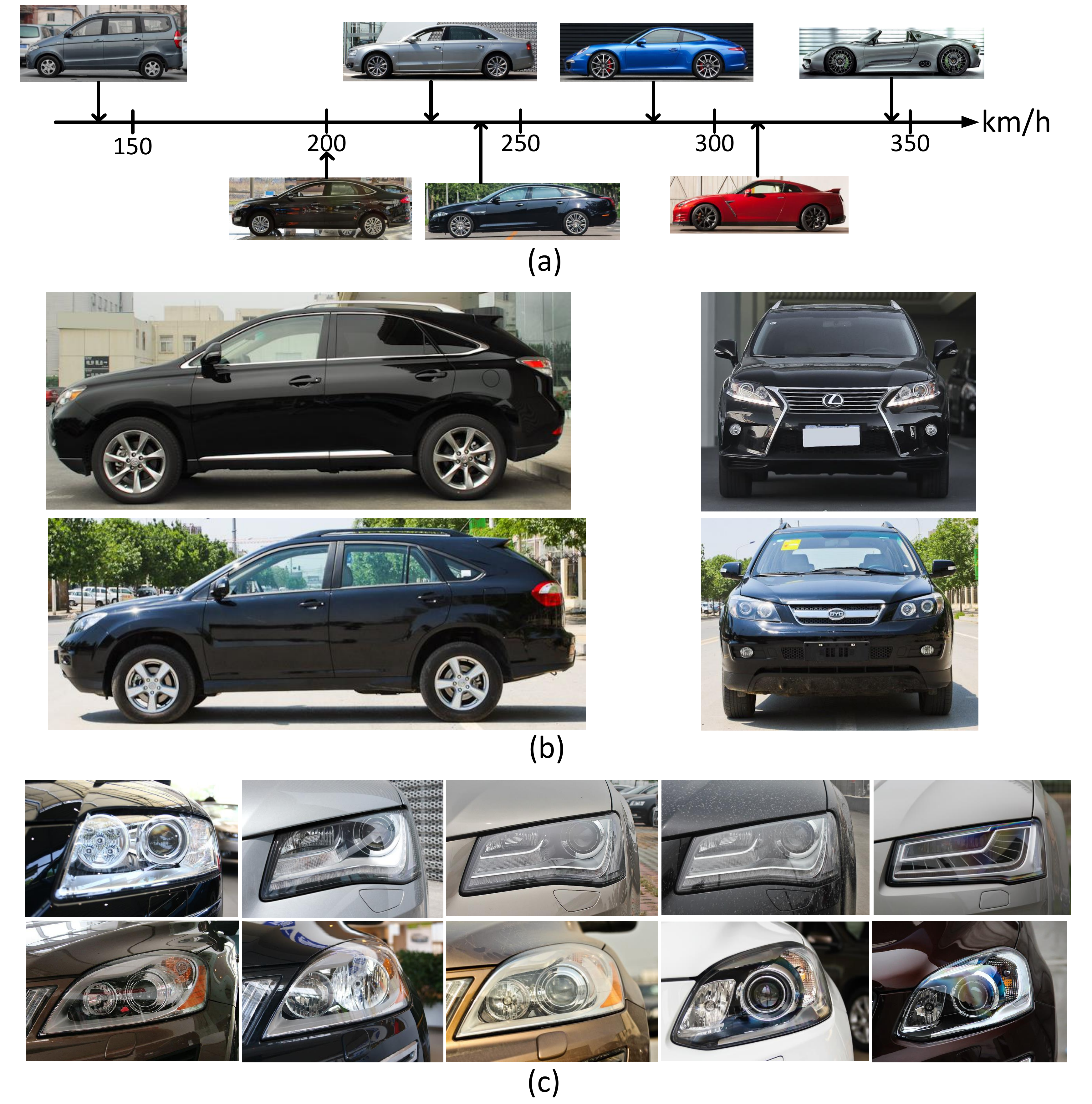}
\caption{(a) Can you predict the maximum speed of a car with only a photo? Get some cues from the examples. (b) The two SUV models are very similar in their side views, but are rather different in the front views. (c) The evolution of the headlights of two car models from 2006 to 2014 (left to right).}
\label{fig:interest}
\vspace{-4pt}
\end{figure}

Cars represent a revolution in mobility and convenience, bringing us the flexibility of moving from place to place. The societal benefits (and cost) are far-reaching.
Cars are now indispensable from our modern life as a vehicle for transportation.
In many places, the car is also viewed as a tool to help project someone's economic status, or reflects our economic stratification. In addition, the car has evolved into a subject of interest amongst many car enthusiasts in the world.
In general, the demand on car has shifted over the years to cover not only practicality and reliability, but also high comfort and design.
The enormous number of car designs and car model makes car a rich object class, which can potentially foster more sophisticated and robust computer vision models and algorithms.

%
%
%




Cars present several unique properties that other objects cannot offer, which provides more challenges and facilitates a range of novel research topics in object categorization.
Specifically, cars own large quantity of models that most other categories do not have, enabling a more challenging fine-grained task.
In addition, cars yield large appearance differences in their unconstrained poses, which demands viewpoint-aware analyses and algorithms (see Fig.~\ref{fig:interest}(b)).
Importantly, a unique hierarchy is presented for the car category, which is three levels from top to bottom: make, model, and released year. This structure indicates a direction to address the fine-grained task in a hierarchical way, which is only discussed by limited literature~\cite{Maji13}.
Apart from the categorization task, cars reveal a number of interesting computer vision problems.
Firstly, different designing styles are applied by different car manufacturers and in different years, which opens the door to fine-grained style analysis~\cite{Lee13style} and fine-grained part recognition (see Fig.~\ref{fig:interest}(c)).
Secondly, the car is an attractive topic for attribute prediction. In particular, cars have distinctive attributes such as car class, seating capacity, number of axles, maximum speed and displacement, which can be inferred from the appearance of the cars (see Fig.~\ref{fig:interest}(a)).
Lastly, in comparison to human face verification~\cite{Sun14}, car verification, which targets at verifying whether two cars belong to the same model, is an interesting and under-researched problem. The unconstrained viewpoints make car verification arguably more challenging than traditional face verification.


Automated car model analysis, particularly the fine-grained car categorization and verification, can be used for innumerable purposes in intelligent transportation system including regulation, description and indexing.
For instance, fine-grained car categorization can be exploited to inexpensively automate and expedite paying tolls from the lanes, based on different rates for different types of vehicles.
%
%
In video surveillance applications, car verification from appearance helps tracking a car over a multiple camera network when car plate recognition fails.
In post-event investigation, similar cars can be retrieved from the database with car verification algorithms.
Car model analysis also bears significant value in the personal car consumption.
When people are planning to buy cars, they tend to observe cars in the street. Think of a mobile application, which can instantly show a user the detailed information of a car once a car photo is taken. Such an application will provide great convenience when people want to know the information of an unrecognized car. Other applications such as predicting popularity based on the appearance of a car, and recommending cars with similar styles can be beneficial both for manufacturers and consumers.



Despite the huge research and practical interests, car model analysis only attracts few attentions in the computer vision community.
We believe the lack of high quality datasets greatly limits the exploration of the community in this domain.
To this end, we collect and organize a large-scale and comprehensive image database called ``Comprehensive Cars'', with ``CompCars'' being short.
The ``CompCars'' dataset is much larger in scale and diversity compared with the current car image datasets, containing $208,826$ images of $1,716$ car models from two scenarios: web-nature and surveillance-nature.
In addition, the dataset is carefully labelled with viewpoints and car parts, as well as rich attributes such as type of car, seat capacity, and door number.
%
%
The new dataset dataset thus provides a comprehensive platform to validate the effectiveness of a wide range of computer vision algorithms. It is also ready to be utilized for realistic applications and enormous novel research topics. Moreover, the multi-scenario nature enables the use of the dataset for cross modality research. The detailed description of \datasetName{} is provided in Section~\ref{sec:dataset}.

To validate the usefulness of the dataset and to encourage the community to explore for more novel research topics, we demonstrate several interesting applications with the dataset, including car model classification and verification based on convolutional neural network (CNN)~\cite{Lecun1989}. Another interesting task is to predict attributes from novel car models (see details in Section~\ref{sec:attr}). The experiments reveal several challenges specific to the car-related problems.
We conclude our analyses with a discussion in Section~\ref{sec:discussion}.

\section{Related Work}


Most previous car model research focuses on car model classification. Zhang \etal~\cite{Zhang12} propose an evolutionary computing framework to fit a wireframe model to the car on an image. Then the wireframe model is employed for car model recognition.
Hsiao \etal~\cite{Hsiao14} construct 3D space curves using 2D training images, then match the 3D curves to 2D image curves using a 3D view-based alignment technique. The car model is finally determined with the alignment result.
Lin \etal~\cite{Lin14} optimize 3D model fitting and fine-grained classification jointly. All these works are restricted to a small number of car models. Recently, Krause et al.~\cite{Krause13} propose to extract 3D car representation for classifying 196 car models. The experiment is the largest scale that we are aware of.
Car model classification is a fine-grained categorization task.
In contrast to general object classification, fine-grained categorization targets at recognizing the subcategories in one object class. Following this line of research, many studies have proposed different datasets on a variety of categories: birds~\cite{CUB_200}, dogs~\cite{Liu12dog}, cars~\cite{Krause13}, flowers~\cite{Nilsback08}, etc. But all these datasets are limited by their scales and subcategory numbers.

To our knowledge, there is no previous attempt on the car model verification task.
Closely related to car model verification, face verification has been a popular topic~\cite{LFWTech,Kumar09,Sun14,ZhuNIPS2014}. The recent deep learning based algorithms~\cite{Sun14} first train a deep neural network on human identity classification, then train a verification model with the feature extracted from the deep neural network. Joint Bayesian~\cite{Chen12} is a widely-used verification model that models two faces jointly with an appropriate prior on the face representation. We adopt Joint Bayesian as a baseline model in car model verification.


Attribute prediction of humans is a popular research topic in recent years~\cite{Bourdev11,DengMM2014,Kumar09,Zhang14}. However, a large portion of the labeled attributes in the current attribute datasets~\cite{DengMM2014}, such as \emph{long hair} and \emph{short pants} lack strict criteria, which causes annotation ambiguities~\cite{Bourdev11}. The attributes with ambiguities will potentially harm the effectiveness of evaluation on related datasets. In contrast, the attributes provided by \datasetName{} (\eg~maximum speed, door number, seat capacity) all have strict criteria since they are set by the car manufacturers. The dataset is thus advantageous over the current datasets in terms of the attributes validity.

Other car-related research includes detection~\cite{Sun06}, tracking~\cite{Matei11}~\cite{Xiang14}, joint detection and pose estimation~\cite{He14, Yang14}, and 3D parsing~\cite{Zia14}. Fine-grained car models are not explored in these studies. Previous research related to car parts includes car logo recognition~\cite{Psyllos10} and car style analysis based on mid-level features~\cite{Lee13style}.

Similar to \datasetName{}, the Cars dataset~\cite{Krause13} also targets at fine-grained tasks on the car category. Apart from the larger-scale database, our \datasetName{} dataset offers several significant benefits in comparison to the Cars dataset.
First, our dataset contains car images diversely distributed in all viewpoints (annotated by front, rear, side, front-side, and rear-side), while Cars dataset mostly consists of front-side car images. Second, our dataset contains aligned car part images, which can be utilized for many computer vision algorithms that demand precise alignment. Third, our dataset provides rich attribute annotations for each car model, which are absent in the Cars dataset.

\section{Properties of \datasetName{}}\label{sec:dataset}

\begin{figure}[t]\centering
\includegraphics[width=0.9\linewidth]{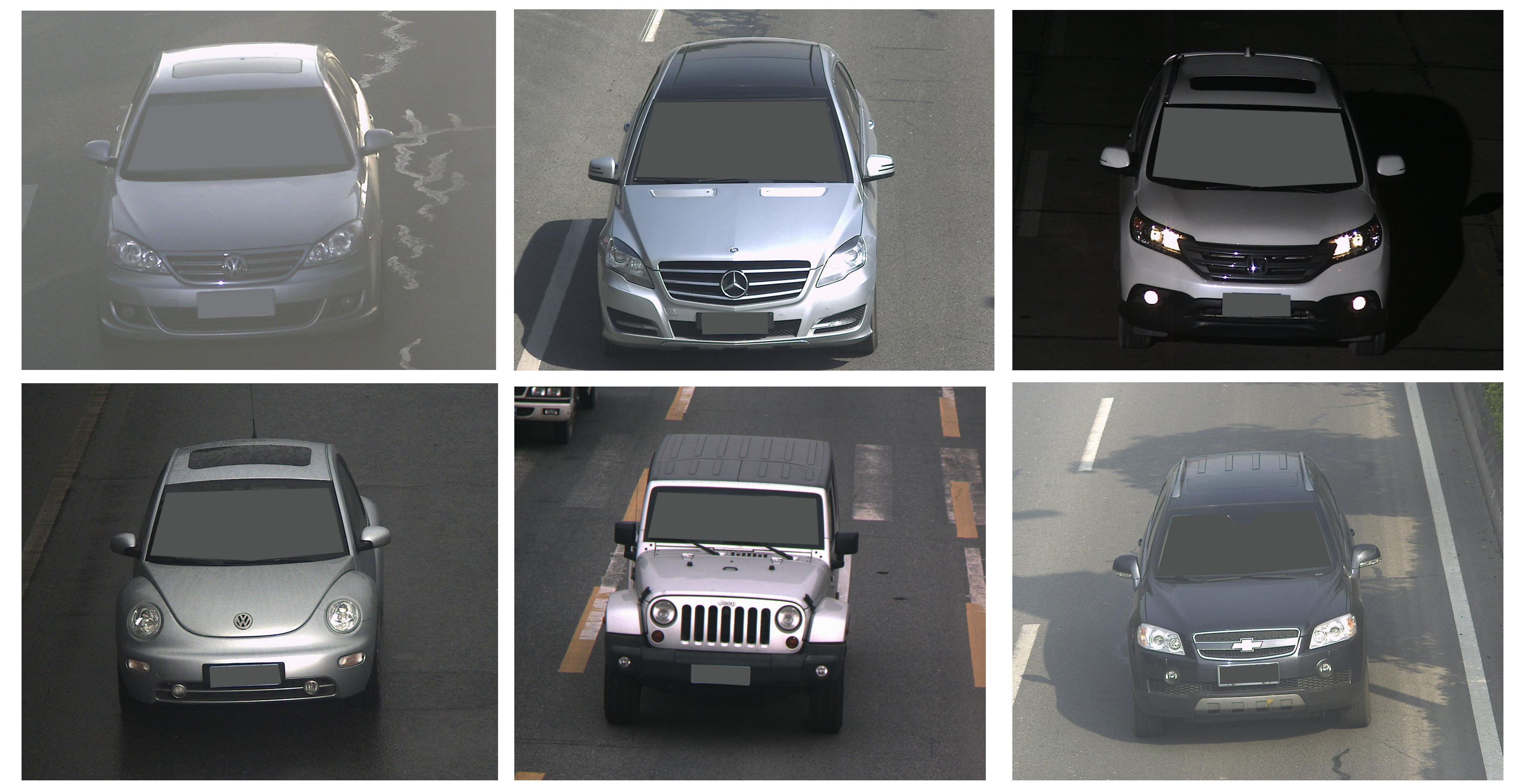}
\caption{Sample images of the surveillance-nature data. The images have large appearance variations due to the varying conditions of light, weather, traffic, etc.}
\label{fig:sv_data}
\vspace{-4pt}
\end{figure}

The \datasetName{} dataset contains data from two scenarios, including images from \emph{web-nature} and \emph{surveillance-nature}.
The images of the web-nature are collected from car forums, public websites, and search engines.
The images of the surveillance-nature are collected by surveillance cameras. The data of these two scenarios are widely used in the real-world applications.
They open the door for cross-modality analysis of cars.
In particular, the web-nature data contains $163$ car makes with $1,716$ car models, covering most of the commercial car models in the recent ten years.
There are a total of $136,727$ images capturing the entire cars and $27,618$ images capturing the car parts, where most of them are labeled with attributes and viewpoints.
%
The surveillance-nature data contains $44,481$ car images captured in the front view.
Each image in the surveillance-nature partition is annotated with bounding box, model, and color of the car.
Fig.~\ref{fig:sv_data} illustrates some examples of surveillance images, which are affected by large variations from lightings and haze.
Note that the data from the surveillance-nature are significantly different from the web-nature data in Fig.~\ref{fig:interest}, suggesting the great challenges in cross-scenario car analysis.
Overall, the \datasetName{} dataset offers four unique features in comparison to existing car image databases, namely car hierarchy, car attributes, viewpoints, and car parts.
%


\begin{figure}[t]\centering
\includegraphics[width=0.8\linewidth]{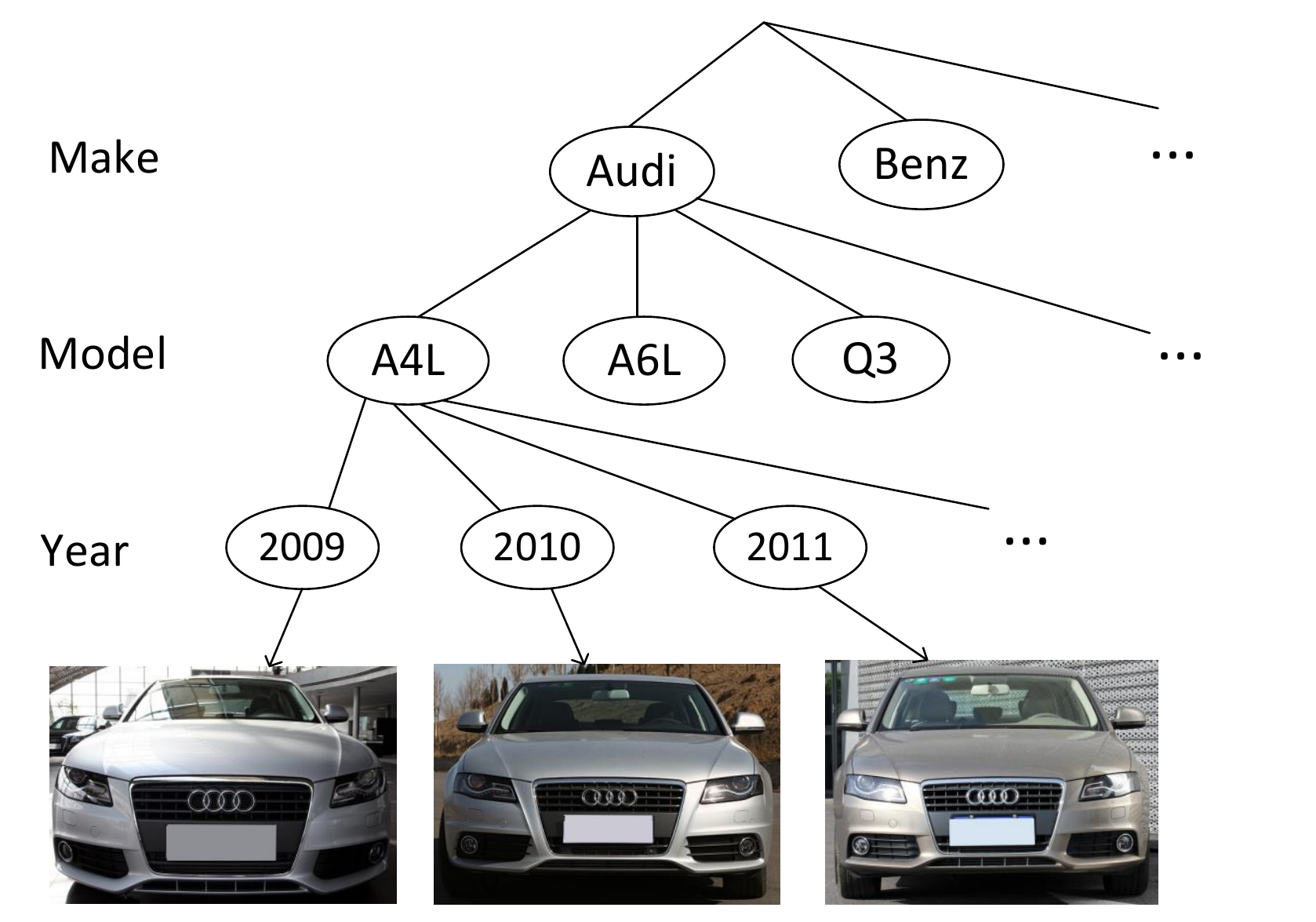}
\caption{The tree structure of car model hierarchy. Several car models of Audi A4L in different years are also displayed. }
\label{fig:tree}
\vspace{-4pt}
\end{figure}

\textbf{Car Hierarchy}
The car models can be organized into a large tree structure, consisting of three layers , namely car make, car model, and year of manufacture, from top to bottom as depicted in Fig.~\ref{fig:tree}.
The complexity is further compounded by the fact that each car model can be produced in different years, yielding subtle difference in their appearances. For instance, three versions of ``Audi A4L'' were produced between 2009 to 2011 respectively.



\begin{figure}[t]\centering
\includegraphics[width=1\linewidth]{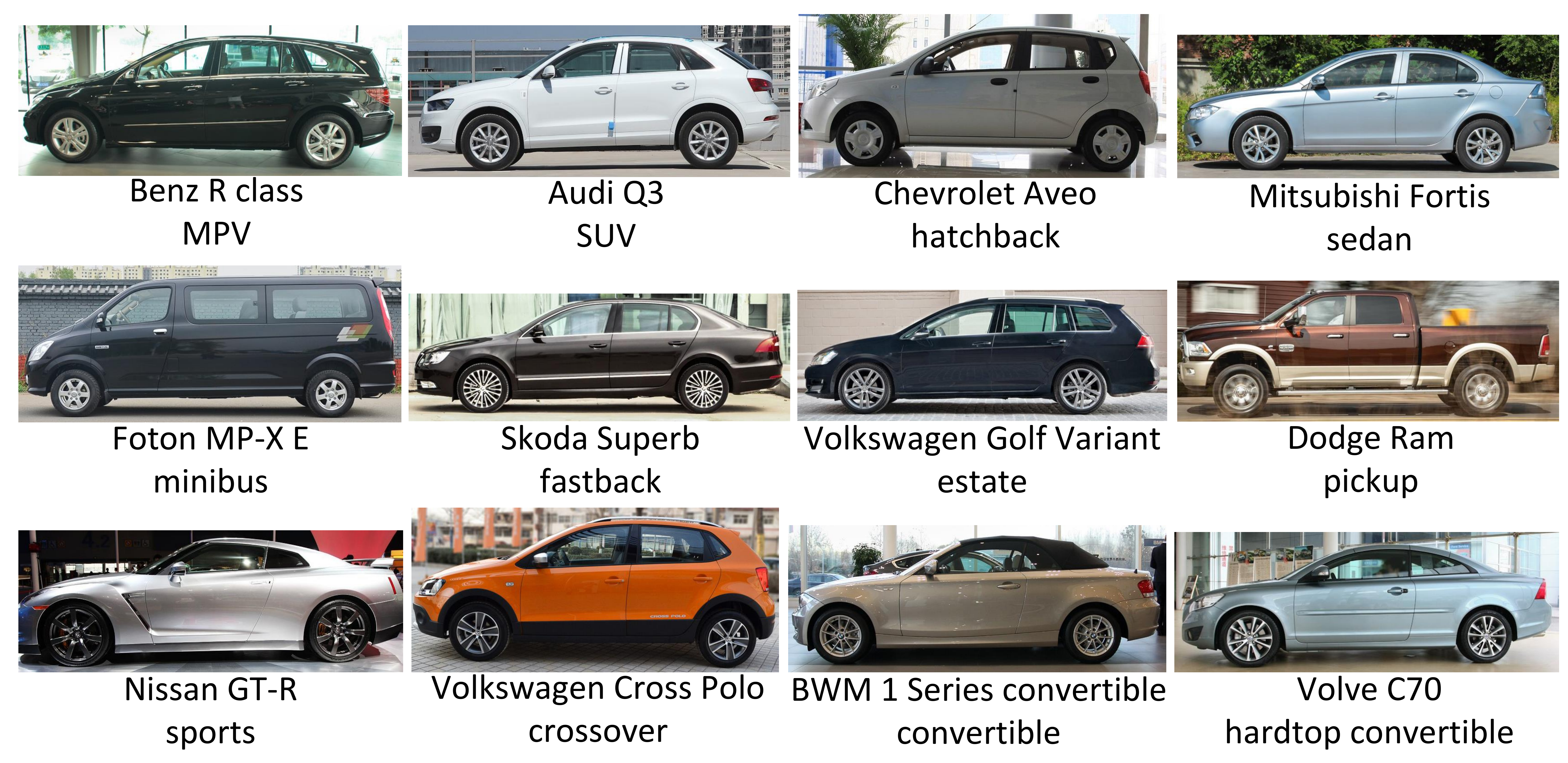}
\caption{Each image displays a car from the 12 car types. The corresponding model names and car types are shown below the images.}
\label{fig:type}
\vspace{-4pt}
\end{figure}

\textbf{Car Attributes}
Each car model is labeled with five attributes, including maximum speed, displacement, number of doors, number of seats, and type of car. These attributes provide rich information while learning the relations or similarities between different car models.
For example, we define twelve types of cars, which are MPV, SUV, hatchback, sedan, minibus, fastback, estate, pickup, sports, crossover, convertible, and hardtop convertible, as shown in Fig.~\ref{fig:type}.
Furthermore, these attributes can be partitioned into two groups: explicit and implicit attributes. The former group contains door number, seat number, and car type, which are represented by discrete values, while the latter group contains maximum speed and displacement (volume of an engine's cylinders), represented by continuous values.
Humans can easily tell the numbers of doors and seats from a car's proper viewpoint, but hardly recognize its maximum speed and displacement.
We conduct interesting experiments to predict these attributes in Section~\ref{sec:attr}.

%
%

\textbf{Viewpoints}
We also label five viewpoints for each car model, including front (F), rear (R), side (S), front-side (FS), and rear-side (RS). These viewpoints are labeled by several professional annotators. The quantity distribution of the labeled car images is shown in Table~\ref{tab:dist_view}.
Note that the numbers of viewpoint images are not balanced among different car models, because the images of some less popular car models are difficult to collect.


%
%


\begin{table}
\small
\centering
\caption{Quantity distribution of the labeled car images in different viewpoints.}
\begin{tabular}{*{3}{|c}|}
\hline
Viewpoint & No. in total  & No. per model \\
\hline
F & 18431 & 10.9\\
\hline
R & 13513 & 8.0\\
\hline
S & 23551 & 14.0\\
\hline
FS & 49301 & 29.2\\
\hline
RS & 31150 & 18.5\\
\hline
\end{tabular}
\label{tab:dist_view}
\vspace{-3pt}
\end{table}

\begin{table}
\small
\centering
\caption{Quantity distribution of the labeled car part images.}
\begin{tabular}{*{3}{|c}|}
\hline
Part & No. in total  & No. per model \\
\hline
headlight & 3705 &	2.2  \\
 \hline
 taillight & 3563 & 2.1 \\
 \hline
 fog light & 3177 &	1.9  \\
 \hline
 air intake & 3407 & 2.0	 \\
 \hline
 console & 3350 &	2.0  \\
 \hline
 steering wheel & 3503 & 2.1 \\
 \hline
 dashboard & 3478 &	2.1  \\
 \hline
 gear lever & 3435 & 2.0 \\
\hline

\end{tabular}
\label{tab:dist_part}
\vspace{-3pt}
\end{table}

\begin{figure}[t]\centering
\includegraphics[width=0.9\linewidth]{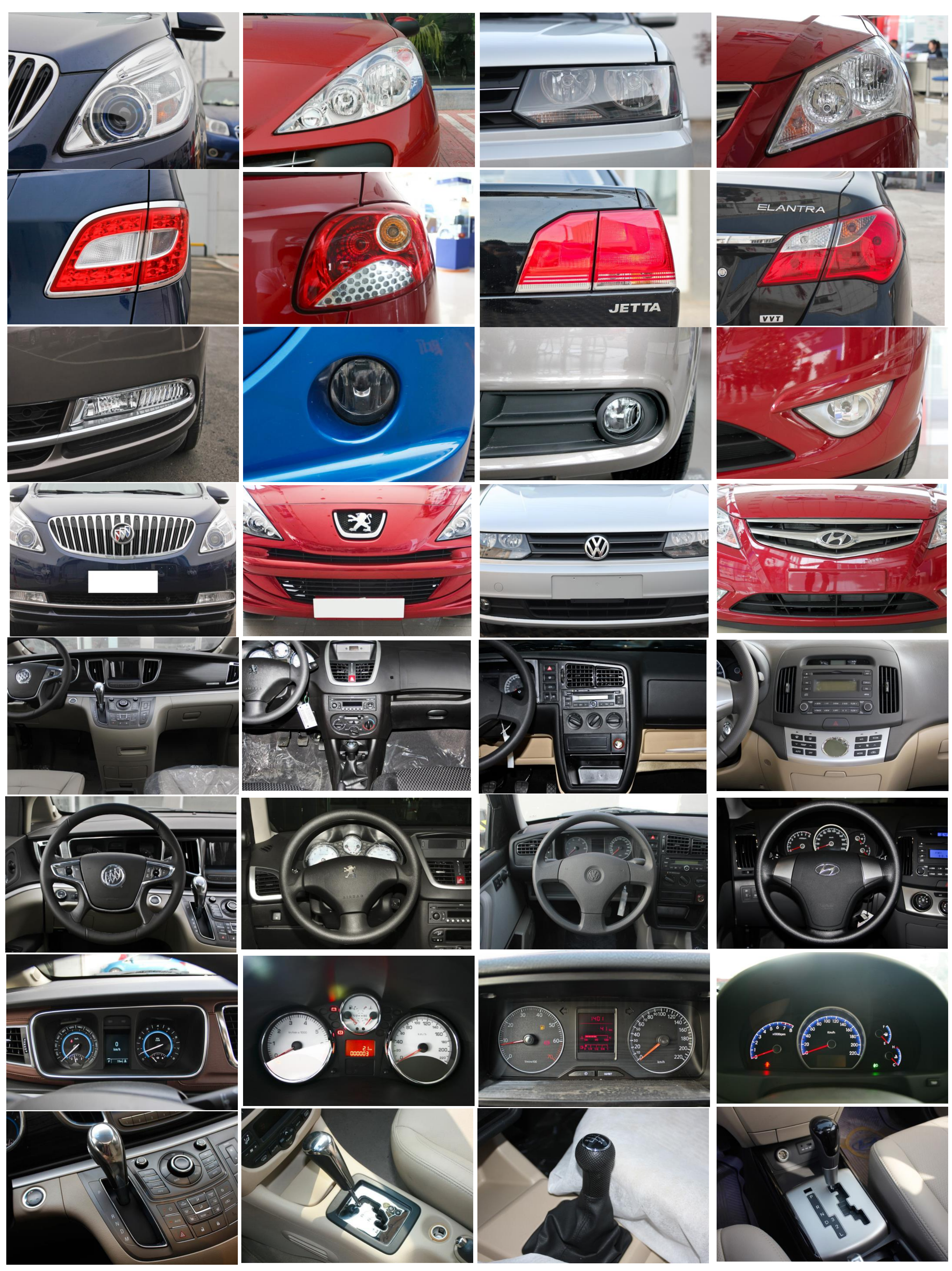}
\caption{Each column displays 8 car parts from a car model. The corresponding car models are Buick GL8, Peugeot 207 hatchback, Volkswagen Jetta, and Hyundai Elantra from left to right, respectively.}
\label{fig:part}
\vspace{-3pt}
\end{figure}

\textbf{Car Parts}
We collect images capturing the eight car parts for each car model, including four exterior parts (\ie headlight, taillight, fog light, and air intake) and four interior parts (\ie console, steering wheel, dashboard, and gear lever). These images are roughly aligned for the convenience of further analysis.
A summary and some examples are given in Table~\ref{tab:dist_part} and Fig.~\ref{fig:part} respectively.

%

\section{Applications}\label{sec:exp}

In this section, we study three applications using \datasetName{}, including fine-grained car classification, attribute prediction, and car verification.
We select $78,126$ images from the \datasetName{} dataset and divide them into three subsets without overlaps. The first subset (Part-I) contains $431$ car models with a total of $30,955$ images capturing the entire car and $20,349$ images capturing car parts. The second subset (Part-II) consists $111$ models with $4,454$ images in total. The last subset (Part-III) contains $1,145$ car models with $22,236$ images.
Fine-grained car classification is conducted using images in the first subset. For attribute prediction, the models are trained on the first subset but tested on the second one. The last subset is utilized for car verification.

We investigate the above potential applications using Convolutional Neural Network (CNN), which achieves great empirical successes in many computer vision problems, such as object classification~\cite{Krizhevsky12}, detection~\cite{rcnn14}, face alignment~\cite{ZhangECCV2014}, and face verification~\cite{Sun14,ZhuNIPS2014}.
Specifically, we employ the Overfeat~\cite{Sermanet13} model, which is pretrained on ImageNet classification task~\cite{Deng09}, and fine-tuned with the car images for car classification and attribute prediction. For car model verification, the fine-tuned model is employed as a feature extractor.



\subsection{Fine-Grained Classification}

We classify the car images into $431$ car models. For each car model, the car images produced in different years are considered as a single category. One may treat them as different categories, leading to a more challenging problem because their differences are relatively small.
%
Our experiments have two settings, comprising fine-grained classification with the entire car images and the car parts. For both settings, we divide the data into half for training and another half for testing.
Car model labels are regarded as training target and logistic loss is used to fine-tune the Overfeat model.

\subsubsection{The Entire Car Images}\label{sec:cls}

\begin{figure}[t]\centering
\includegraphics[width=1\linewidth]{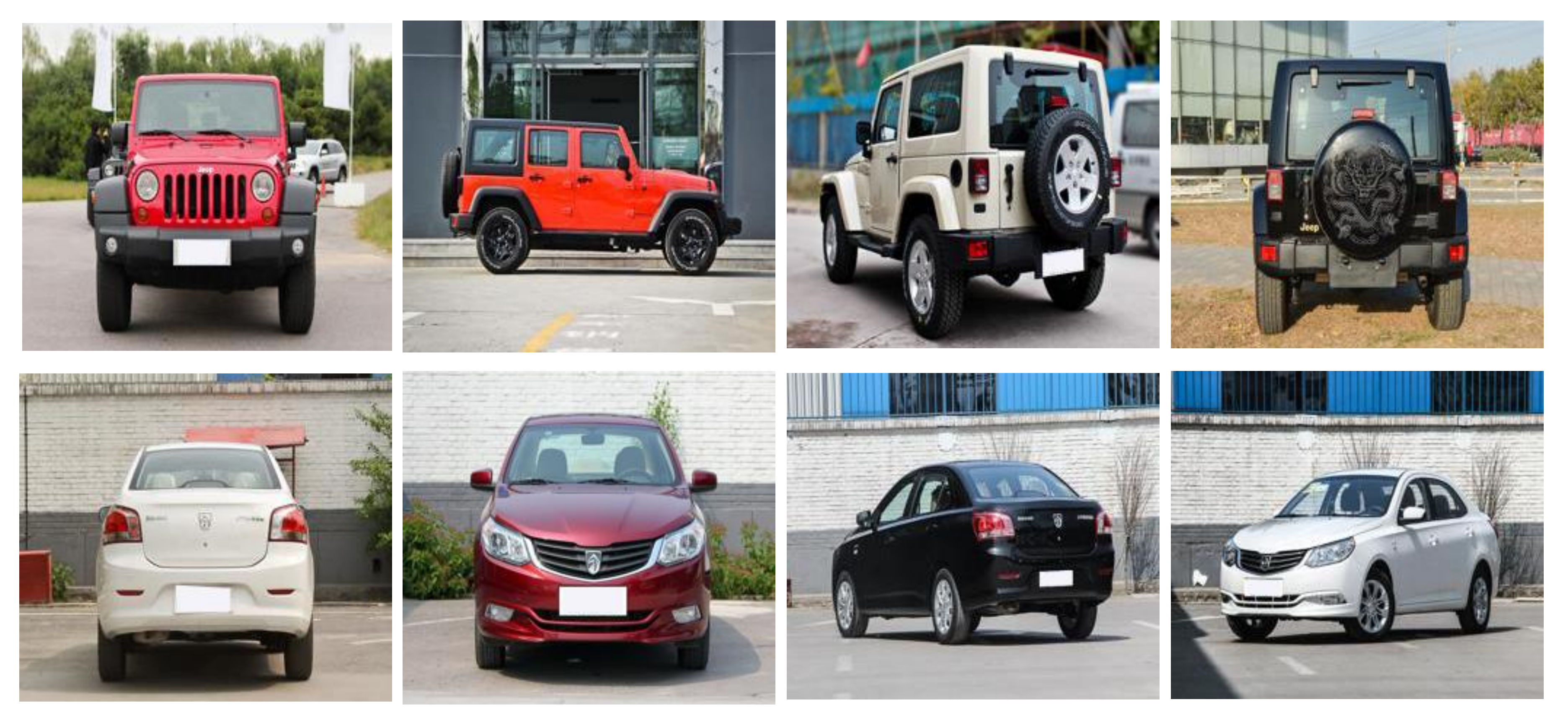}
\caption{Images with the highest responses from two sample neurons. Each row corresponds to a neuron.}
\label{fig:response}
\vskip -0.35cm
\end{figure}

We compare the recognition performances of the CNN models, which are fine-tuned with car images in specific viewpoints and all the viewpoints respectively, denoted as ``front (F)'', ``rear (R)'', ``side (S)'', ``front-side (FS)'', ``rear-side (RS)'', and ``All-View''.
The performances of these six models are summarized in Table~\ref{tab:cls_car}, where ``FS'' and ``RS'' achieve better performances than the performances of the other viewpoint models.
Surprisingly, the ``All-View'' model yields the best performance, although it did not leverage the information of viewpoints. This result reveals that the CNN model is capable of learning discriminative representation across different views.
To verify this observation, we visualize the car images that trigger high responses with respect to each neuron in the last fully-connected layer. As shown in Fig.~\ref{fig:response}, these neurons capture car images of specific car models across different viewpoints.

\begin{figure}[t]\centering
\includegraphics[width=1\linewidth]{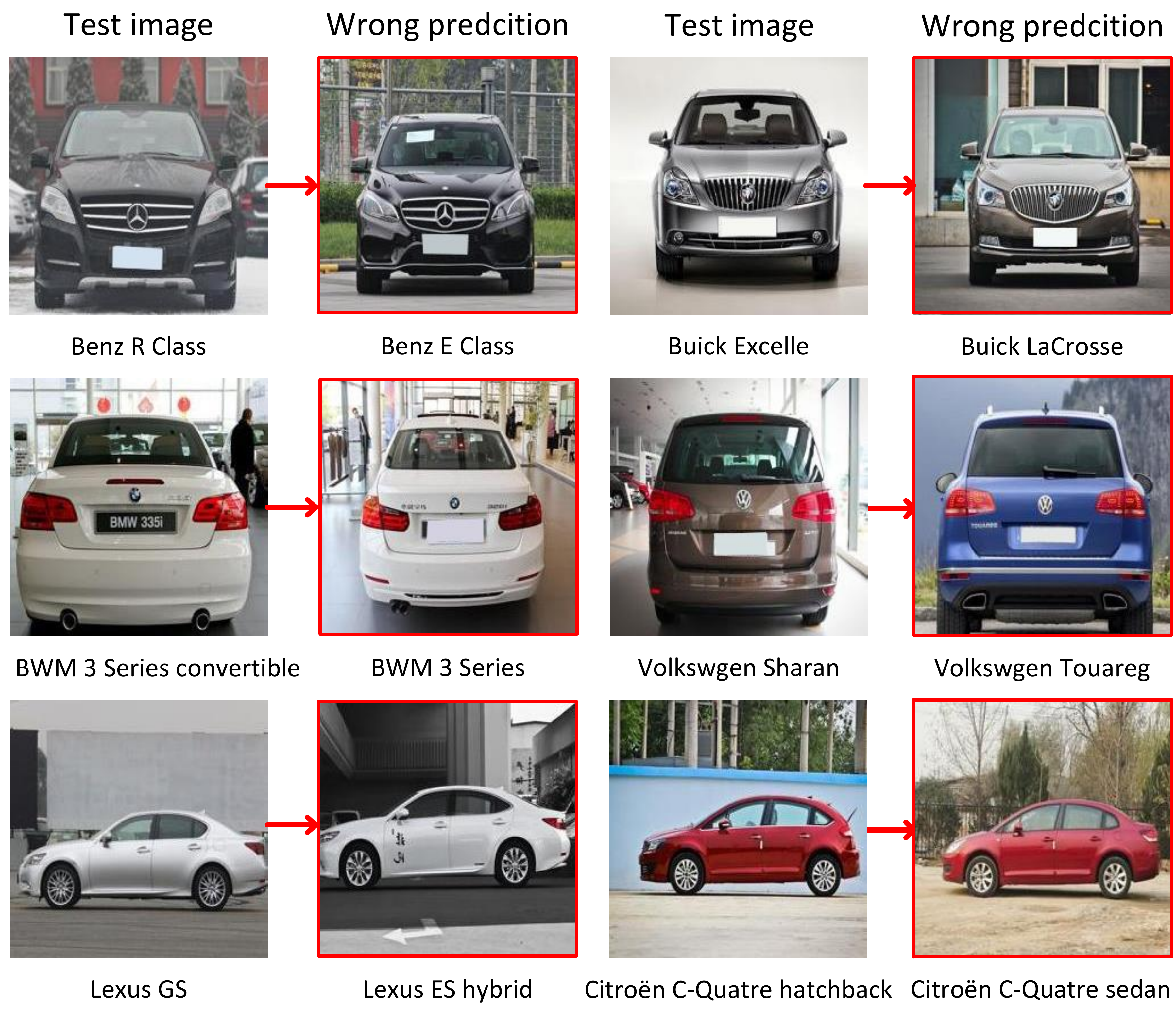}
\caption{Sample test images that are mistakenly predicted as another model in their makes. Each row displays two samples and each sample is a test image followed by another image showing its mistakenly predicted model. The corresponding model name is shown under each image.}
\label{fig:cls_err}
\vspace{-3pt}
\end{figure}

\begin{figure}[t]\centering
\includegraphics[width=1.0\linewidth]{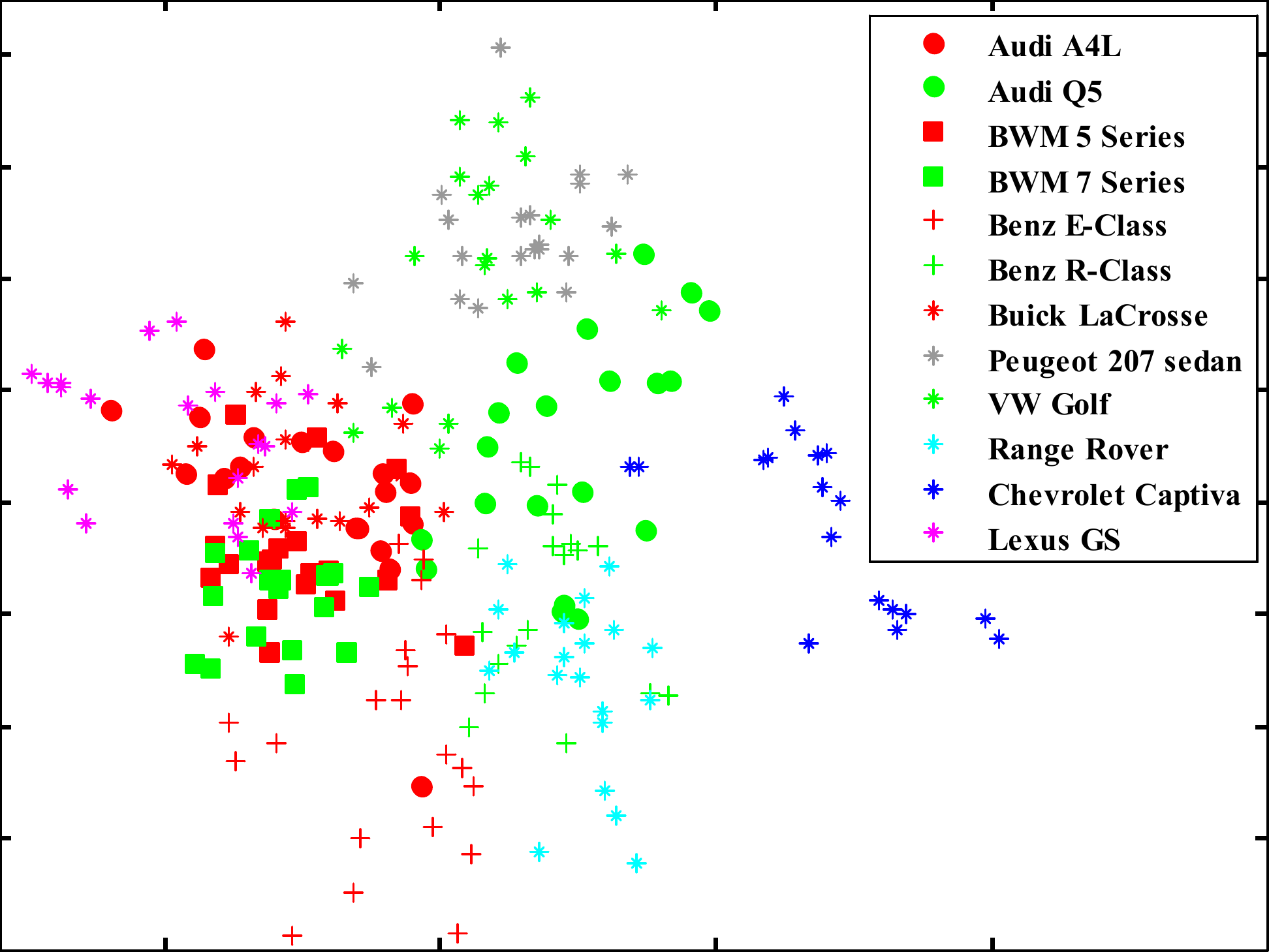}
\caption{The features of 12 car models that are projected to a two-dimensional embedding using multi-dimensional scaling. Most features are separated in the 2D plane with regard to different models. Features extracted from similar models such as BWM 5 Series and BWM 7 Series are close to each other. Best viewed in color.}
\label{fig:dist}
\vskip -0.35cm
\end{figure}

Several challenging cases are given in Fig.~\ref{fig:cls_err}, where the images on the left hand side are the testing images and the images on the right hand side are the examples of the wrong predictions (of the ``All-View'' model). We found that most of the wrong predictions belong to the same car makes as the test images.
%
%
We report the ``top-1'' accuracies of car make classification in the last row of Table~\ref{tab:cls_car}, where the ``All-View'' model obtain reasonable good result, indicating that a coarse-to-fine (\ie from car make to model) classification is possible for fine-grained car recognition.


To observe the learned feature space of the ``All-View'' model, we project the features extracted from the last fully-connected layer to a two-dimensional embedding space using multi-dimensional scaling. Fig.~\ref{fig:dist} visualizes the projected features of twelve car models, where the images are chosen from different viewpoints.
%
We observe that features from different models are separable in the 2D space and features of similar models are closer than those of dissimilar models. For instance, the distances between ``BWM 5 Series'' and ``BWM 7 Series'' are smaller than those between ``BWM 5 Series'' and ``Chevrolet Captiva''. 
\begin{table}
\small
\centering
\caption{Fine-grained classification results for the models trained on car images. Top-1 and Top-5 denote the top-1 and top-5 accuracy for car model classification, respectively. Make denotes the make level classification accuracy.}
\begin{tabular}{*{7}{|c}|}
\hline
 Viewpoint & F & R & S & FS & RS & All-View\\
\hline
Top-1 & 0.524 &	0.431 &	0.428 & 0.563 & 0.598 &	 0.767\\
Top-5 & 0.748 & 0.647 & 0.602 & 0.769 & 0.777 &  0.917\\
\hline
Make & 0.710 & 0.521 & 0.507 & 0.680 & 0.656 &  0.829\\
\hline
\end{tabular}
\label{tab:cls_car}
\vspace{-3pt}
\end{table}

\begin{figure}[t]\centering
\includegraphics[width=1\linewidth]{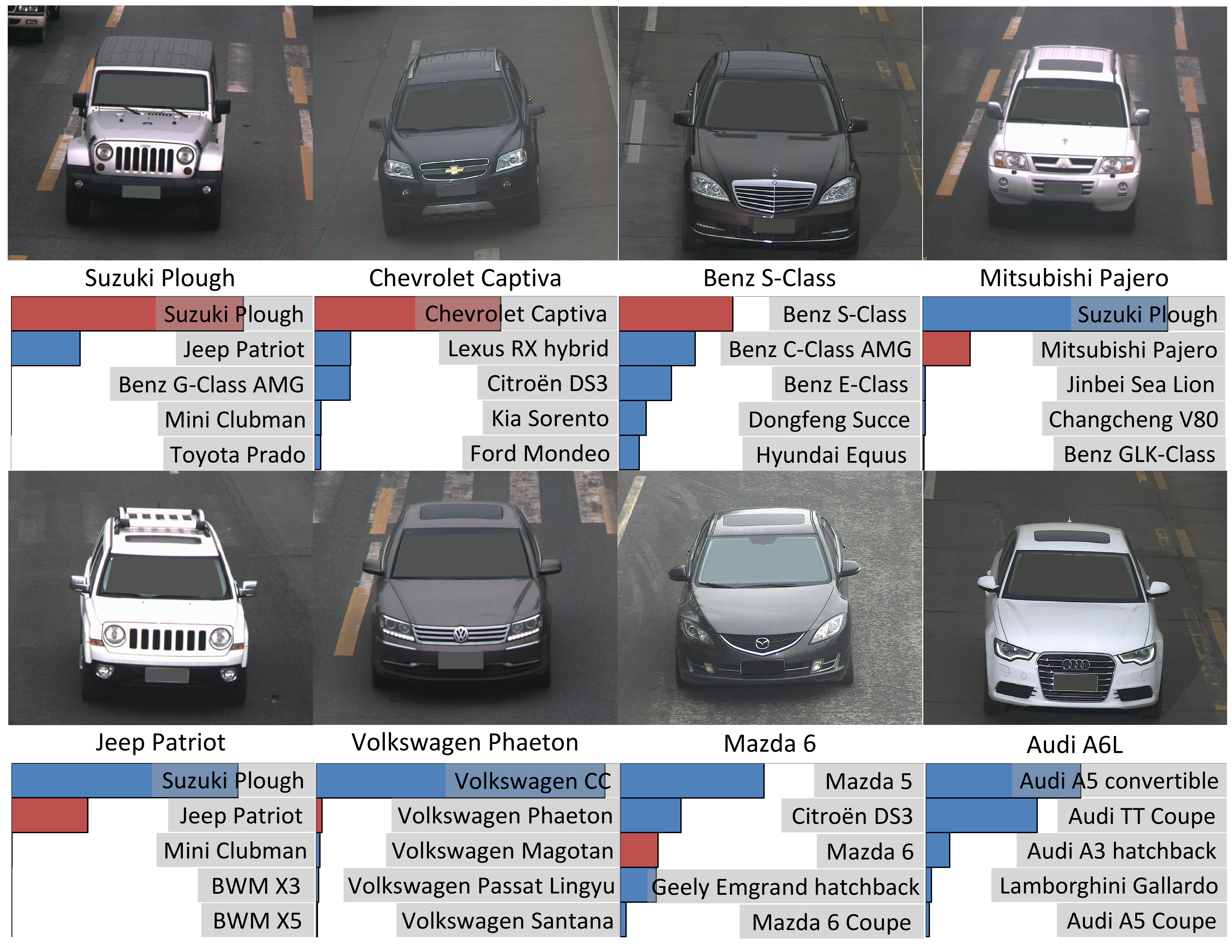}
\caption{Top-5 predicted classes of the classification model for eight cars in the surveillance-nature data. Below each image is the ground truth class and the probabilities for the top-5 predictions with the correct class labeled in red. Best viewed in color.}
\label{fig:cross_t5}
\vspace{-3pt}
\end{figure}

We also conduct a cross-modality experiment, where the CNN model fine-tuned by the web-nature data is evaluated on the surveillance-nature data. Fig.~\ref{fig:cross_t5} illustrates some predictions, suggesting that the model may account for data variations in a different modality to a certain extent.
This experiment indicates that the features obtained from the web-nature data have potential to be transferred to data in the other scenario.


\subsubsection{Car Parts}

\begin{figure}[t]\centering
\includegraphics[width=0.7\linewidth]{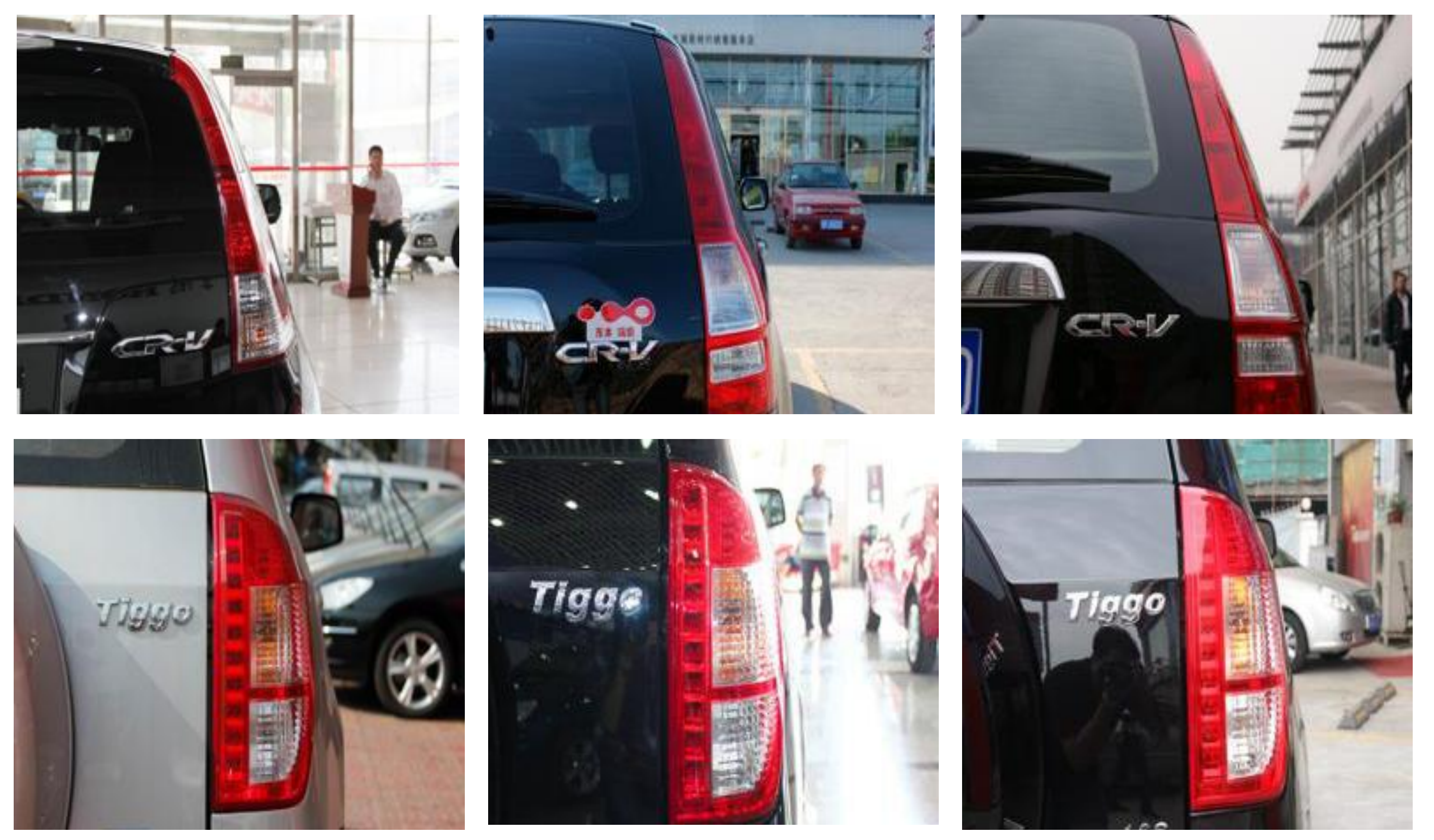}
\caption{Taillight images with the highest responses from two sample neurons. Each row corresponds to a neuron.}
\label{fig:part_response}
\end{figure}

Car enthusiasts are able to distinguish car models by examining the car parts. We investigate if the CNN model can mimic this strength.
We train a CNN model using images from each of the eight car parts. The results are reported in Table~\ref{tab:cls_part}, where ``taillight'' demonstrates the best accuracy.
We visualize taillight images that have high responses with respect to each neuron in the last fully-connected layer. Fig.~\ref{fig:part_response} displays such images with respect to two neurons. ``Taillight'' wins among the different car parts, mostly likely due to the relatively more distinctive designs, and the model name printed close to the taillight, which is a very informative feature for the CNN model.


We also combine predictions using the eight car part models by voting strategy. This strategy significantly improves the performance due to the complementary nature of different car parts.

\begin{table*}
\small
\centering
\caption{Fine-grained classification results for the models trained on car parts. Top-1 and Top-5 denote the top-1 and top-5 accuracy for car model classification, respectively.}
\begin{tabular}{*{10}{|c}|}
\hline
& \multicolumn{4}{c|}{Exterior parts} & \multicolumn{4}{c|}{Interior parts} & \\
\cline{2-9}
  & Headlight & Taillight & Fog light & Air intake & Console & Steering wheel & Dashboard & Gear lever & Voting\\
\hline
Top-1 & 0.479 &0.684& 0.387& 0.484& 0.535& 0.540& 0.502& 0.355& 0.808 \\
\hline
Top-5 & 0.690 &0.859& 0.566 & 0.695& 0.745& 0.773& 0.736& 0.589& 0.927 \\
\hline
\end{tabular}
\label{tab:cls_part}
\vspace{-3pt}
\end{table*}


\subsection{Attribute Prediction}\label{sec:attr}

\begin{figure}[t]\centering
\includegraphics[width=0.9\linewidth]{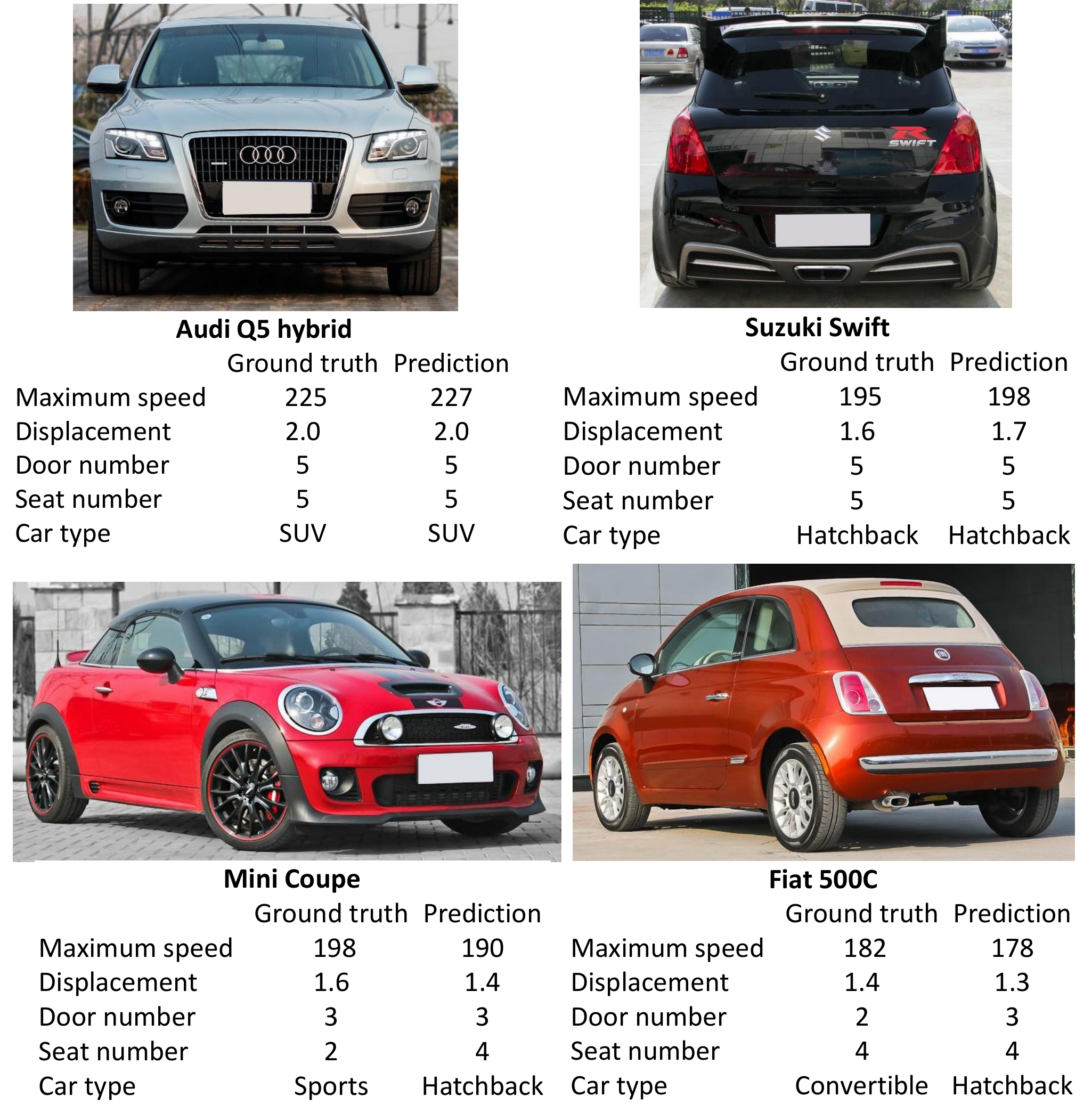}
\caption{Sample attribute predictions for four car images. The continuous predictions of maximum speed and displacement are rounded to nearest proper values.}
\label{fig:attr}
\vspace{-3pt}
\end{figure}

Human can easily identify the car attributes such as numbers of doors and seats from a proper viewpoint, without knowing the car model.
For example, a car image captured in the side view provides sufficient information of the door number and car type, but it is hard to infer these attributes from the frontal view.
The appearance of a car also provides hints on the implicit attributes, such as the maximum speed and the displacement. For instance, a car model is probably designed for high-speed driving, if it has a low under-pan and a streamline body.

In this section, we deliberately design a challenging experimental setting for attribute recognition, where the car models presented in the test images are exclusive from the training images.
We fine-tune the CNN with the sum-of-square loss to model the continuous attributes, such as ``maximum speed'' and ``displacement'', but a logistic loss to predict the discrete attributes such as ``door number'', ``seat number'', and ``car type''. For example, the ``door number'' has four states, \ie~$\left\{2, 3, 4, 5\right\}$ doors, while ``seat number'' also has four states, \ie~$\left\{2, 4, 5, >5\right\}$ seats. The attribute ``car type'' has twelve states as discussed in Sec.~\ref{sec:dataset}.




To study the effectiveness of different viewpoints for attribute prediction, we train CNN models for different viewpoints separately.
Table~\ref{tab:attr_car} summarizes the results, where the ``mean guess'' represents the errors computed by using the mean of the training set as the prediction.
We observe that the performances of ``maximum speed'' and ``displacement'' are insensitive to viewpoints. However, for the explicit attributes, the best accuracy is obtained under side view.
%
%
We also found that the the implicit attributes are more difficult to predict then the explicit attributes. Several test images and their attribute predictions are provided in Fig.~\ref{fig:attr}. 




 \begin{table}
\small
\centering
\caption{Attribute prediction results for the five single viewpoint models. For the continuous attributes (maximum speed and displacement), we display the mean difference from the ground truth. For the discrete attributes (door and seat number, car type), we display the classification accuracy. Mean guess denotes the mean error with a prediction of the mean value on the training set.}
\begin{tabular}{*{6}{|c}|}
\hline
 Viewpoint & F & R & S & FS & RS  \\
\hline
& \multicolumn{5}{c|}{mean difference} \\
\hline
Maximum speed & 20.8 & 21.3 & 20.4 & 20.1 & 21.3 	 \\
(mean guess) & 38.0 &38.5 &39.4 & 40.2 & 40.1 \\
Displacement & 0.811 & 0.752 & 0.795 & 0.875 & 0.822   \\
(mean guess) & 1.04 & 0.922 & 1.04 & 1.13 & 1.08 \\
\hline
& \multicolumn{5}{c|}{classification accuracy} \\
\hline
Door number &0.674 & 0.748& 0.837& 0.738& 0.788 \\
Seat number &0.672 &0.691 &0.711 &0.660 &0.700 \\
Car type & 0.541 &0.585 &0.627 &0.571 &0.612  \\
\hline
\end{tabular}
\label{tab:attr_car}
\vspace{-3pt}
\end{table}

\subsection{Car Verification}\label{sec:verif}
In this section, we perform car verification following the pipeline of face verification~\cite{Sun14}.
In particular, we adopt the classification model in Section~\ref{sec:cls} as a feature extractor of the car images, and then apply Joint Bayesian~\cite{Chen12} to train a verification model on the Part-II data. Finally, we test the performance of the model on the Part-III data, which includes $1,145$ car models.
The test data is organized into three sets, each of which has different difficulty, \ie easy, medium, and hard. Each set contains $20,000$ pairs of images, including $10,000$ positive pairs and $10,000$ negative pairs.
Each image pair in the ``easy set'' is selected from the same viewpoint, while each pair in the ``medium set'' is selected from a pair of random viewpoints. Each negative pair in the ``hard set'' is chosen from the same car make.
 %

Deeply learned feature combined with Joint Bayesian has been proven successful for face verification~\cite{Sun14}. Joint Bayesian formulates the feature $x$ as the sum of two independent Gaussian variables
\begin{align}
  x=\mu + \epsilon,
\end{align}
where $\mu \sim N(0,S_{\mu})$ represents identity information, and $\epsilon \sim N(0,S_{\epsilon})$ the intra-category variations. Joint Bayesian models the joint probability of two objects given the intra or extra-category variation hypothesis, $P(x_1,x_2| H_I)$ and $P(x_1, x_2 | H_E)$. These two probabilities are also Gaussian with variations
\begin{align}
  \Sigma_I = \left[ \begin{array}{cc}
  S_{\mu}+S_{\epsilon} & S_{\mu}\\
  S_{\mu} & S_{\mu} + S_{\epsilon} \end{array} \right]
\end{align}
and
\begin{align}
  \Sigma_E = \left[ \begin{array}{cc}
  S_{\mu}+S_{\epsilon} &0\\
  0 & S_{\mu} + S_{\epsilon}
  \end{array} \right],
\end{align}
respectively. $S_{\mu}$ and $S_{\epsilon}$ can be learned from data with EM algorithm. In the testing stage, it calculates the likelihood ratio
\begin{align}
  r(x_1, x_2) = \log \frac{P(x_1, x_2 | H_I)}{P(x_1, x_2 | H_E)},
\end{align}
which has closed-form solution.
The feature extracted from the CNN model has a dimension of $4,096$, which is reduced to $20$ by PCA. The compressed features are then utilized to train the Joint Bayesian model.
%
During the testing stage, each image pair is classified by comparing the likelihood ratio produced by Joint Bayesian with a threshold. This model is denoted as (CNN feature + Joint Bayesian).

The second method combines the CNN features and SVM, denoted as CNN feature + SVM. Here, SVM is a binary classifier using a pair of image features as input. The label `1' represents positive pair, while `0' represents negative pair.
We extract $100,000$ pairs of image features from Part-II data for training.

\begin{figure}[t]\centering
\includegraphics[width=1\linewidth]{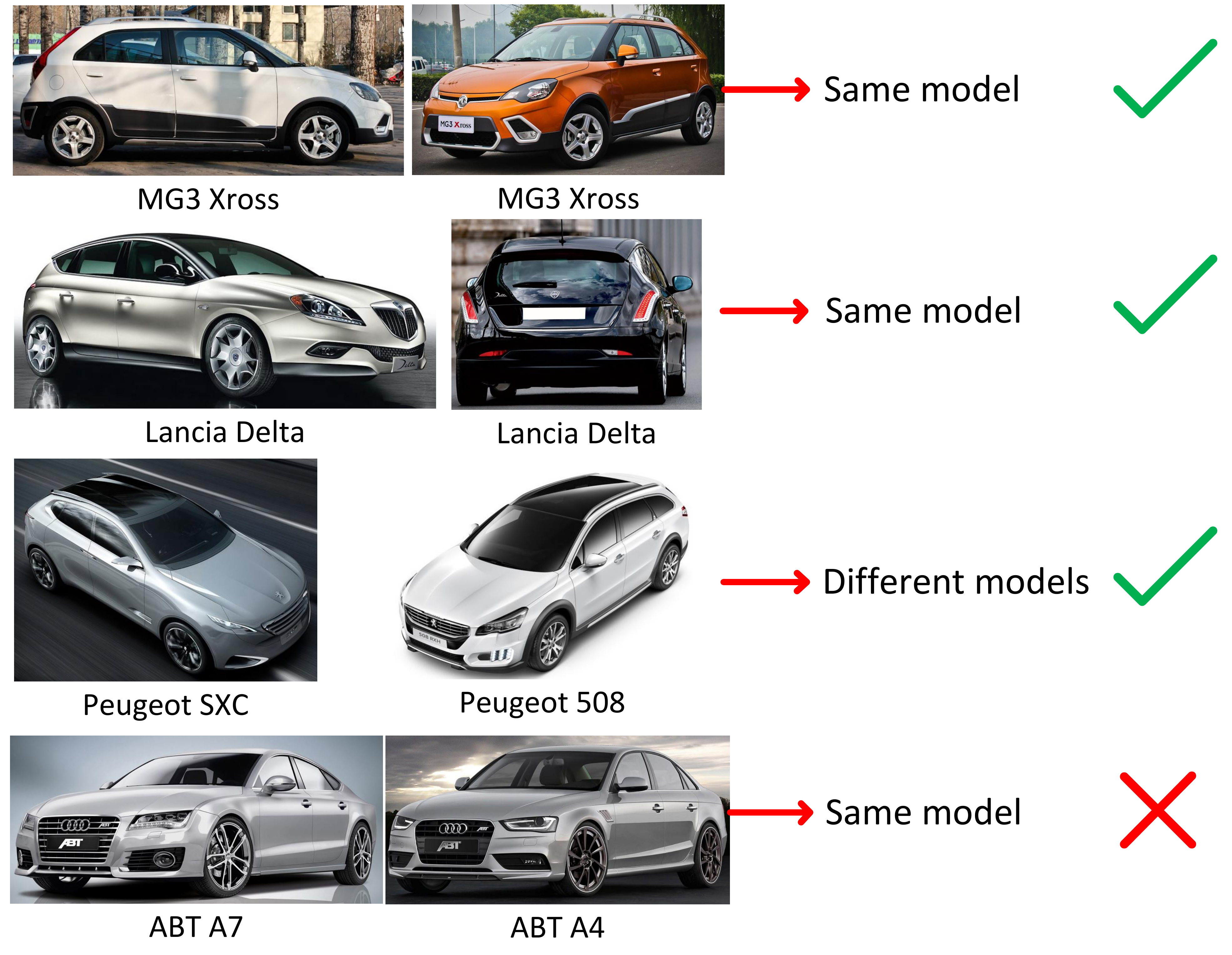}
\caption{Four test samples of verification and their prediction results. All these samples are very challenging and our model obtains correct results except for the last one. }
\label{fig:verif}
\vspace{-3pt}
\end{figure}

The performances of the two models are shown in Table~\ref{tab:verif} and the ROC curves for the ``hard set'' are plotted in Fig.~\ref{fig:roc}.
We observe that CNN feature + Joint Bayesian outperforms CNN feature + SVM with large margins, indicating the advantage of Joint Bayesian for this task.
However, its benefit in car verification is not as effective as in face verification, where CNN and Joint Bayesian nearly saturated the LFW dataset \cite{LFWTech} and approached human performance~\cite{Sun14}.
Fig.~\ref{fig:verif} depicts several pairs of test images as well as their predictions by CNN feature + Joint Bayesian. We observe two major challenges. First, for the image pair of the same model but different viewpoints, it is difficult to obtain the correspondences directly from the raw image pixels. Second, the appearances of different car models of the same car make are extremely similar. It is difficult to distinguish these car models using the entire images. Part localization or detection is crucial for car verification.

%

\begin{table}
\small
\centering
\caption{The verification accuracy of three baseline models.  }
\begin{tabular}{*{4}{|c}|}
\hline
   & Easy & Medium& Hard\\
\hline
CNN feature + Joint Bayesian  & 0.833& 0.824& 0.761 \\
\hline
CNN feature + SVM & 0.700 & 0.690 & 0.659 \\
\hline
random guess & \multicolumn{3}{c|}{0.500} \\
\hline
\end{tabular}
\label{tab:verif}
\end{table}

\begin{figure}[t]\centering
\includegraphics[width=1\linewidth]{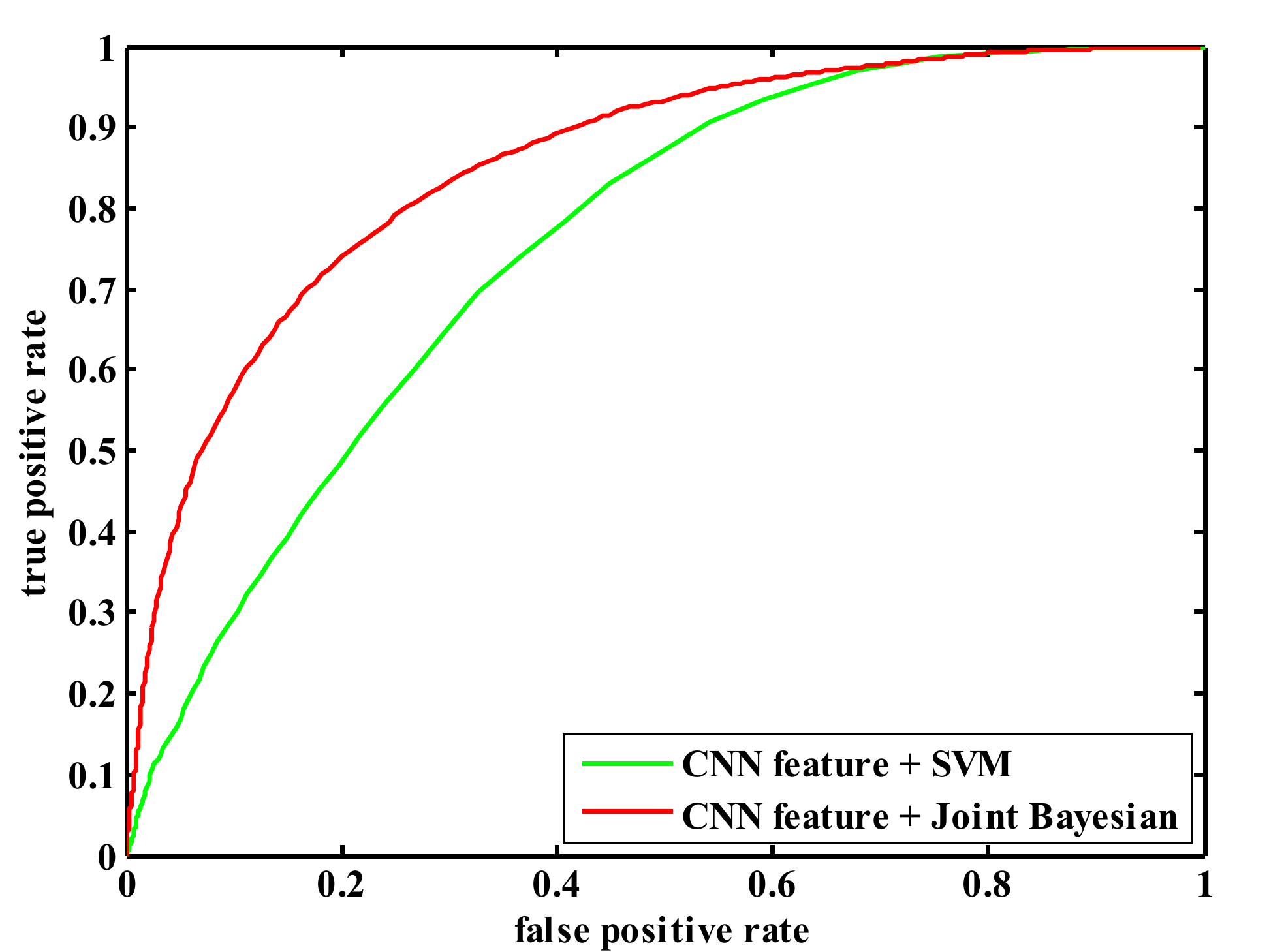}
\caption{The ROC curves of two baseline models for the hard flavor. }
\label{fig:roc}
\end{figure}

\section{Updated Results: Comparing Different Deep Models}\label{sec:updated}
As an extension to the experiments in Section~\ref{sec:exp}, we conduct experiments for fine-grained car classification, attribute prediction, and car verification with the entire dataset and different deep models, in order to explore the different capabilities of the models on these tasks. The split of the dataset into the three tasks is similar to Section~\ref{sec:exp}, where three subsets contain $431$, $111$, and $1,145$ car models, with $52,083$, $11,129$, and $72,962$ images respectively. The only difference is that we adopt full set of CompCars in order to establish updated baseline experiments and to make use of the dataset to the largest extent. We keep the testing sets of car verification same to those in Section~\ref{sec:verif}.

We evaluate three network structures, namely AlexNet~\cite{Krizhevsky12}, Overfeat~\cite{Sermanet13}, and GoogLeNet~\cite{Szegedy14} for all three tasks. All networks are pre-trained on the ImageNet classification task~\cite{Deng09}, and fine-tuned with the same mini-batch size, epochs, and learning rates for each task. All predictions of the deep models are produced with a single center crop of the image. We use Caffe~\cite{Jia14caffe} as the platform for our experiments. The experimental results can serve as baselines in any later research works. The train/test splits can be downloaded from CompCars webpage \url{http://mmlab.ie.cuhk.edu.hk/datasets/comp_cars/index.html}.

\subsection{Fine-Grained Classification}\label{sec:class}
In this section, we classify the car images into $431$ car models as in Section~\ref{sec:cls}. We divide the data into 70\% for training and 30\% for testing. We train classification models using car images in all viewpoints. The performances of the three networks are summarized in Table~\ref{tab:class}. Overfeat beats AlexNet with a large margin of 6.0\% while GoogLeNet beats Overfeat by 3.3\% in Top-1 accuracy, which is in consistency with their performances on the ImageNet classification task. Given more data, the accuracy rises about 11\% for Overfeat compared to Table~\ref{tab:cls_car}\footnote{Due to the difference in testing sets, the accuracies are not directly comparable. However a rough estimate is still viable.}. We also release the fine-tuned GoogLeNet model on the CompCars webpage.

\begin{table}
\small
\centering
\caption{The classification accuracies of three deep models.}
\begin{tabular}{*{4}{|c}|}
\hline
Model & AlexNet  & Overfeat & GoogLeNet \\
\hline
Top-1 & 0.819 &	0.879 & 0.912 \\
 \hline
Top-5 & 0.940 & 0.969 & 0.981 \\
 \hline

\end{tabular}
\label{tab:class}
\end{table}

\subsection{Attribute Prediction}\label{sec:attr_up}
We predict attributes from $111$ models not existed in the training set. Different from Section~\ref{sec:attr} where models are trained with cars in single viewpoints, we train with images in all viewpoints to build a compact model. Table~\ref{tab:attr_up} summarizes the results for the three networks, where ``mean guess'' represents the prediction with the mean of the values on the training set. GoogLeNet performs the best for all attributes and Overfeat is a close running-up.

\begin{table}
\small
\centering
\caption{Attribute prediction results of three deep models. For the continuous attributes (maximum speed and displacement), we display the mean difference from the ground truth (lower is better). For the discrete attributes (door and seat number, car type), we display the classification accuracy (higher is better).}
\begin{tabular}{*{4}{|c}|}
\hline
Model & AlexNet  & Overfeat & GoogLeNet \\
\hline
& \multicolumn{3}{c|}{mean difference} \\
\hline
Maximum speed & 21.3 &	19.4 & 19.4  \\
\cline{2-4}
(mean guess) &\multicolumn{3}{c|}{36.9 } \\
\cline{2-4}
Displacement & 0.803 & 0.770 & 0.760  \\
\cline{2-4}
(mean guess) &\multicolumn{3}{c|}{1.02 } \\
 \hline
& \multicolumn{3}{c|}{classification accuracy} \\
\hline
Door number & 0.750 &0.780 & 0.796 \\
Seat number & 0.691&0.713 &0.717  \\
Car type & 0.602& 0.631 & 0.643 \\
\hline
\end{tabular}
\label{tab:attr_up}
\end{table}

\subsection{Car Verification}
The evaluation pipeline follows Section~\ref{sec:verif}. We evaluate the three deep models combined with two verification models: Joint Bayesian~\cite{Chen12} and SVM with polynomial kernel. The feature extracted from the CNN models is reduced to $200$ by PCA before training and testing in all experiments.

The performances of the three networks combined with the two verification models are shown in Table~\ref{tab:verif_up}, where each model is denoted by \{name of the deep model\} + \{name of the verification model\}. GoogLeNet + Joint Bayesian achieves the best performance in all three settings. For each deep model, Joint Bayesian outperforms SVM consistently. Compared to Table~\ref{tab:verif}, Overfeat + Joint Bayesian yields a performance gain of $2\sim4\%$ in the three settings, which is purely due to the increase in training data.
The ROC curves for the three sets are plotted in Figure~\ref{fig:roc}.

\begin{table}
\small
\centering
\caption{The verification accuracies of six models.}
\begin{tabular}{*{4}{|c}|}
\hline
  & Easy  & Medium & Hard \\
\hline
AlexNet + SVM & 0.822 &	0.800 & 0.729 \\
 \hline
AlexNet + Joint Bayesian & 0.853 &	0.823 & 0.774 \\
 \hline
Overfeat + SVM & 0.860 & 0.830 & 0.754 \\
 \hline
Overfeat + Joint Bayesian & 0.873 &	0.841 & 0.780 \\
 \hline
GoogLeNet + SVM & 0.880 &	0.837 & 0.764 \\
 \hline
GoogLeNet + Joint Bayesian & 0.907 & 0.852 & 0.788 \\
 \hline
\end{tabular}
\label{tab:verif_up}
\end{table}

\begin{figure}[t]
\small\centering
\begin{minipage}[t]{0.9\linewidth}
\centering
\includegraphics[height=2.2in]{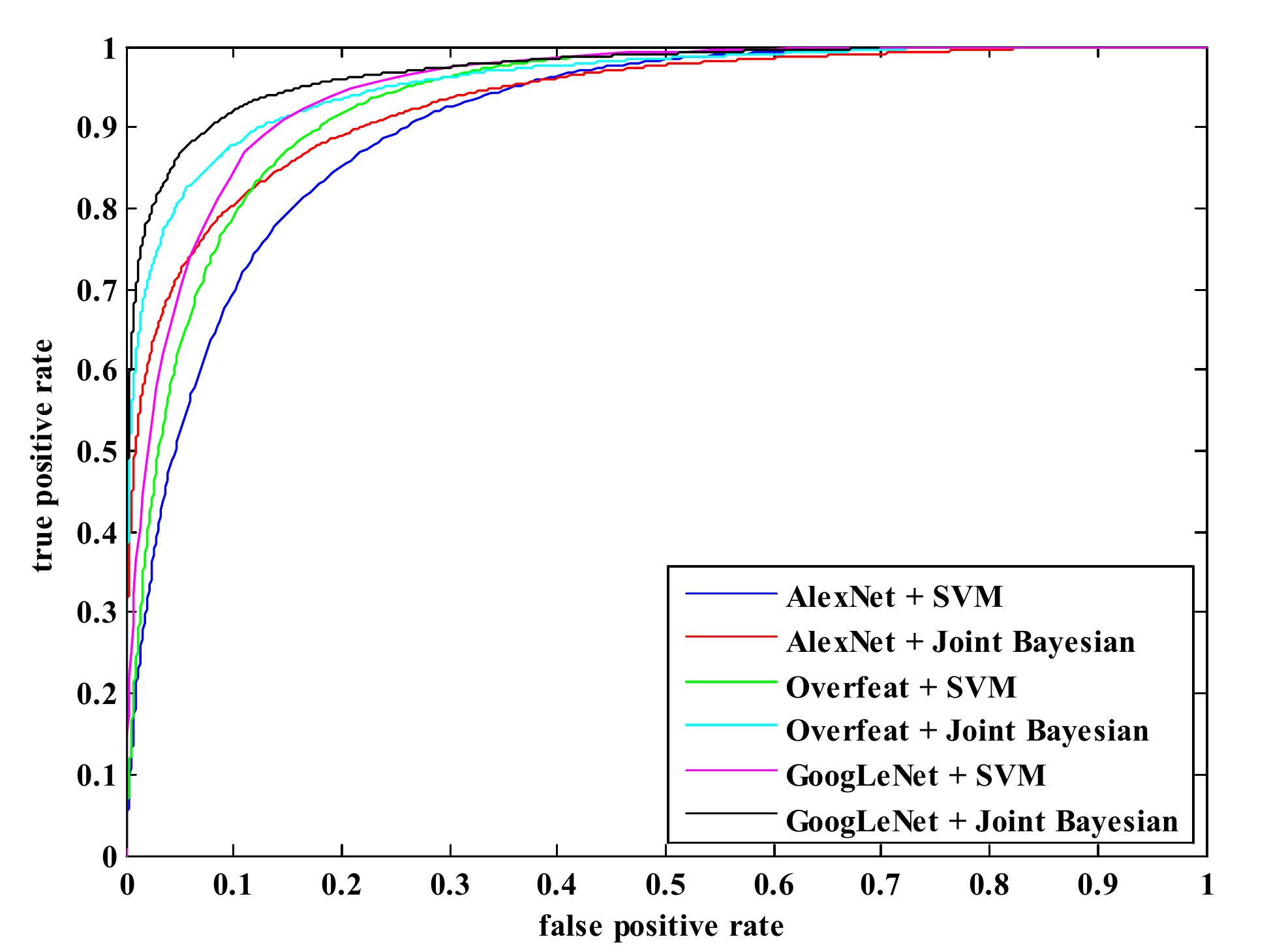}
\par{(a)}
\end{minipage}
\begin{minipage}[t]{0.9\linewidth}
\centering
\includegraphics[height=2.2in]{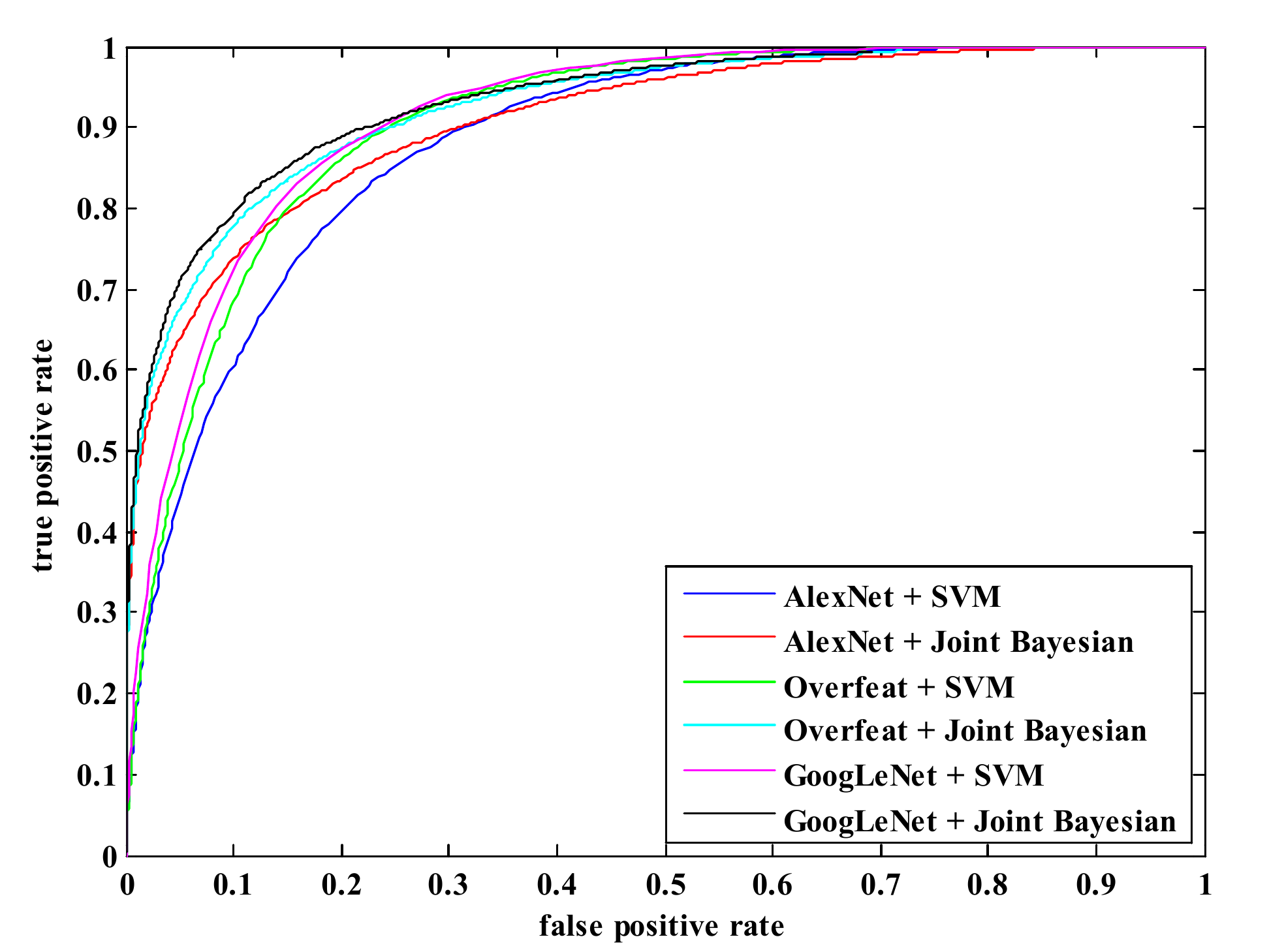}
\par{(b)}
\end{minipage}
\begin{minipage}[t]{0.9\linewidth}
\centering
\includegraphics[height=2.2in]{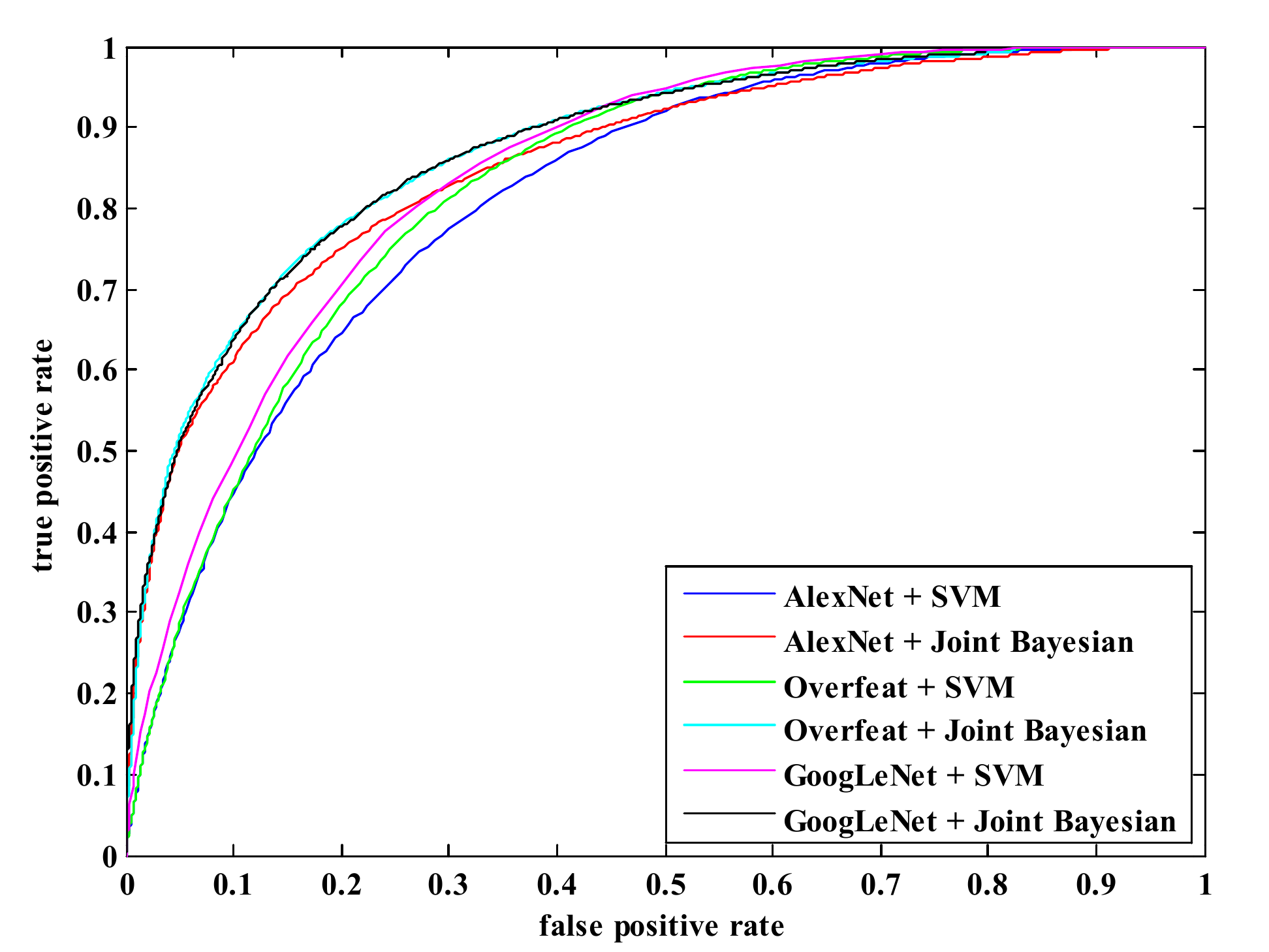}
\par{(c)}
\end{minipage}
\caption{The ROC curves of six verification models for (a) easy, (b) medium, and (c) hard set. }
\label{fig:roc}
\end{figure}

\section{Fine-Grained Classification with Surveillance Data}\label{sec:surveillance}
This is a follow-up experiment for fine-grained classification with surveillance-nature data. The data includes $44,481$ images in $281$ different car models. 70\% images are for training and 30\% are for testing. The car images are all in front views with various environment conditions such as rainy, foggy, and at night. We adopt the same three network structures (AlexNet, Overfeat, and GoogLeNet) as in the web-nature data applications for this task. The networks are also pre-trained on the ImageNet classification task, and the test is done with a single center crop. The car images are first cropped with the labeled bounding boxes with paddings of around 7\% on each side. All cropped images are resized to $256\times256$ pixels. The experimental results are shown in Table~\ref{tab:surveillance}. The three networks all achieve very high accuracies for this task. The result indicates that the fixed view (front view) greatly simplifies the fine-grained classification task, even when large environmental differences exist.

\begin{table}
\small
\centering
\caption{The classification accuracies of three deep models on surveillance data.}
\begin{tabular}{*{4}{|c}|}
\hline
Model & AlexNet  & Overfeat & GoogLeNet \\
\hline
Top-1 & 0.980 &	0.983 & 0.984 \\
 \hline
\end{tabular}
\label{tab:surveillance}
\end{table}

\section{Discussions}\label{sec:discussion}

In this paper, we wish to promote the field of research related to ``cars'', which is largely neglected by the computer vision community. To this end, we have introduced a large-scale car dataset called \datasetName{}, which contains images with not only different viewpoints, but also car parts and rich attributes. \datasetName{} provides a number of unique properties that other fine-grained datasets do not have, such as a much larger subcategory quantity, a unique hierarchical structure, implicit and explicit attributes, and large amount of car part images which can be utilized for style analysis and part recognition. It also bears cross modality nature, consisting of web-nature data and surveillance-nature data, ready to be used for cross modality research.
To validate the usefulness of the dataset and inspire the community for other novel tasks, we have conducted baseline experiments on three tasks: car model classification, car model verification, and attribute prediction. The experimental results reveal several challenges of these tasks and provide qualitative observations of the data, which is beneficial for future research.

There are many other potential tasks that can exploit \datasetName{}. Image ranking is one of the long-lasting topics in the literature, car model ranking can be adapted from this line of research to find the models that users are mostly interested in. The rich attributes of the dataset can be used to learn the relationships between different car models. Combining with the provided 3-level hierarchy, it will yield a stronger and more meaningful relationship graph for car models. Car images from different viewpoints can be utilized for ultra-wide baseline matching and 3D reconstruction, which can benefit recognition and verification in return.

{\small
\bibliographystyle{ieee}
\bibliography{egbib}
}

\end{document}